\documentclass[sigconf]{acmart}

\AtBeginDocument{%
  }

\title{Embedding Knowledge Graph in Function Space}

\copyrightyear{2024}
\acmYear{2024}
\setcopyright{rightsretained}
\acmConference[CIKM '24]{Proceedings of the 33rd ACM International
Conference on Information and Knowledge Management}{October 21--25,
2024}{Boise, ID, USA}
\acmBooktitle{Proceedings of the 33rd ACM International Conference on
Information and Knowledge Management (CIKM '24), October 21--25, 2024,
Boise, ID, USA}
\acmDOI{10.1145/3627673.3679819}
\acmISBN{979-8-4007-0436-9/24/10}
\author{Louis Mozart Kamdem Teyou}
\email{louis888@mail.uni-paderborn.de}
\orcid{0000-0001-7975-8794}
\affiliation{%
  \institution{Paderborn University}
   \department{Electronic, Informatic and Mathematic}
  \city{Paderborn}
  \state{North Rhine-Westphalia}
  \country{Germany}
}

\author{Caglar Demir}
\email{caglar.demir@uni-paderborn.de}
\orcid{0000-0001-8970-3850}
\affiliation{%
  \institution{Paderborn University}
  \department{Electronic, Informatic and Mathematic}
  \city{Paderborn}
  \state{North Rhine-Westphalia}
  \country{Germany}
}

\author{Axel-Cyrille Ngonga Ngomo}
\orcid{0000-0001-7112-3516}
\email{axel.ngonga@upb.de}
\affiliation{%
  \institution{Paderborn University}
   \department{Electronic, Informatic and Mathematic}
  \city{Paderborn}
  \state{North Rhine-Westphalia}
  \country{Germany}
}

\usepackage{xspace}
\usepackage{graphicx}
\usepackage{multirow} 
\usepackage{paralist}
\usepackage{subcaption}
\usepackage{balance}

\title{Embedding Knowledge Graphs in Function Spaces}
\author{}
\DeclareGraphicsExtensions{.pdf,.png,.jpg}

\makeatletter
\gdef\@copyrightpermission{
  \begin{minipage}{0.3\columnwidth}
   \href{https://creativecommons.org/licenses/by-nc-sa/4.0/}{\includegraphics[width=0.90\textwidth]{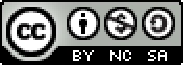}}
  \end{minipage}\hfill
  \begin{minipage}{0.7\columnwidth}
   \href{https://creativecommons.org/licenses/by-nc-sa/4.0/}{This work is licensed under a Creative Commons Attribution-NonCommercial-ShareAlike International 4.0 License.}
  \end{minipage}
  \vspace{5pt}
}
\makeatother

\begin{document}

% \maketitle

 \newcommand{\fspace}{\ensuremath{\mathcal{F}}\xspace}
\newcommand{\an}[1]{\todo[color=yellow]{AN: #1}}
\newcommand{\triple}[3]{\ensuremath{\langle \texttt{#1}, \texttt{#2}, \texttt{#3} \rangle}}
\newcommand{\approach}{\textsc{FMult}\xspace} 
\newcommand{\KG}{\mathcal{K}\xspace} 

\newcommand{\keci}{\textsc{Keci}\xspace}

\begin{abstract}
  We introduce a novel embedding method diverging from conventional approaches by operating within function spaces of finite dimension rather than finite vector space, thus departing significantly from standard knowledge graph embedding techniques. Initially employing polynomial functions to compute embeddings, we progress to more intricate representations using neural networks with varying layer complexities. We argue that employing functions for embedding computation enhances expressiveness and allows for more degrees of freedom, enabling operations such as composition, derivatives and primitive of entities representation. Additionally, we meticulously outline the step-by-step construction of our approach and provide code for reproducibility, thereby facilitating further exploration and application in the field.
\end{abstract}




% \ccsdesc[500]{Computing methodologies~Reasoning about belief and knowledge}
% \ccsdesc[500]{Computing methodologies~Ranking}



\maketitle

\section{Introduction}

A knowledge graph (KG) serves as a structured representation of data, aiming to encapsulate and convey real-world knowledge \cite{DBLP:journals/csur/HoganBCdMGKGNNN21}. Comprising triples that articulate facts about the world, KG data are pivotal, in organizing information. For example, the triple $\langle Obama, president\_of, America \rangle$ represents the fact that Barack Obama was the president of the United States. Consequently, KGs find applications in various domains such as web search \cite{Singhal2012}, recommender systems \cite{wang2019kgat, li2023survey}, and natural language processing \cite{nickel2011three}.

Given that KGs are essentially composed of strings or characters, knowledge graph embedding (KGE) involves mapping entities and relations from a KG into a vector space \cite{DBLP:journals/tkde/WangMWG17}. This transformation facilitates computational operations, enabling the application of machine learning and deep learning techniques to extract insights from the KG. Hence, an effective KGE model should strive to maintain the properties and semantics inherent in the original KG.

Knowledge graphs are typically represented as $K \subseteq \mathcal{E} \times \mathcal{R} \times \mathcal{E}$, where $\mathcal{E}$ and $\mathcal{R}$ represent sets of entities and relations respectively. They are commonly embedded  in $d$-dimensional vector spaces $\mathbb{V}$ such as $\mathbb{R}^d$ (real numbers), $\mathbb{C}^d$ (complex numbers), or even $\mathbb{H}^d$ (quaternions) \cite{bordes2013translating, trouillon2016complex, chami2020low}. While such embeddings offer a low-dimensional representation, they treat entities and relations as static vectors, which may limit their ability to capture some dynamics in the knowledge graph.  For instance, in real-world applications, relationships between entities can change over time exemplifying temporal dynamics (e.g.  transition from $"is\_friends\_with"$ to $"was\_friends\_with."$) \cite{jain2022representation, leblay2020towards}. They can also depend on context, such as in movie recommendations, where the relevance of a movie may be context-dependent (e.g., $"is\_interested\_in\_watching"$ may vary based on user preferences  \cite{cao2021dual}. Static embeddings may fail to represent such dynamic behaviours accurately, leading to suboptimal performance in tasks like link prediction or knowledge graph completion \cite{kazemi2018simple}.

% \sh{I would probably write the next line in a separate paragraph along with the next paragraph.}

To mitigate the aforementioned drawback of the existing embedding approaches, in this work, we propose the use of functions to represent entities and relations of a KG, i.e., embedding them in function space.
%Alternative- Functional representations of ..
Functional representations of entities and relations provide a compelling alternative to static embeddings. For instance, functions can be time-dependent, they can support compositionality, enabling the combination of simpler functions to represent more complex relationships and entities. 
%Alternative- Moreover, functions offer a richer and more expressive representation compared to static embeddings, allowing the modeling of complex interactions and dependencies within KGs.
Moreover, functions offer a richer and more expressive representation compared to  vectors, allowing the modelling of complex interactions and dependencies within KGs. Also, functions offer greater interpretability, as they can be analyzed and understood based on their mathematical properties and behaviour. By examining the functional forms, it would be possible to gain insights into the underlying structure of the knowledge representation, thereby facilitating better understanding and interpretation.

Building upon these advantages, we hypothesize that embedding KGs in a function space offers increased degrees of freedom in the embeddings.
Therefore, in this paper, we leverage the advantages of \textbf{f}unction \textbf{mult}iplication and composition to compute the embeddings in function space. Our methods, dubbed $\approach{}_n$, $\approach{}^i_n$, and $\approach{}$, compute embeddings by representing entities and relations as functions. $\approach{}_n$ employs polynomials to model the interactions between entities and relations, capturing non-linear relationships that static vectors might miss. $\approach{}^i_n$ utilizes trigonometric functions to incorporate periodic patterns and cyclic behaviours inherent in the KG. Finally, $\approach{}$ leverages neural networks to learn complex, high-dimensional embeddings that can adapt dynamically to various contexts.

% \sh{maybe two or three sentences regarding the approach using these functions?}

This paradigm shift allows for a more flexible and expressive representation of entities and relations. Notably, to the best of our knowledge, our model is the first link prediction model to use functions for representing entities and relations in KGs. Our results have shown better performance compared to traditional static embedding methods, highlighting the effectiveness of functional embeddings in capturing the complex dynamics of KGs.

In summary, we make the following contributions in this work:
\begin{itemize}
\item We propose three different functional embedding approaches $\approach{}_n$, $\approach{}^i_n$, and $\approach{}$ to embed the entities and relations of KGs using polynomials, trigonometric functions, and neural networks, respectively.
\item We perform a comprehensive evaluation of our proposed methods on multiple benchmark datasets, demonstrating their superior performance in link prediction tasks compared to state-of-the-art models.
\item We make our implementations and experimental results publicly available to facilitate further research and reproducibility in the field. \footnote{https://github.com/dice-group/dice-embeddings}
\end{itemize}
The remainder of the paper is organized as follows: In the next section, we present related work in the KGE field. In Sections \ref{sec:Function space} and \ref{sec:Methodology}, we provide preliminaries on function space and show how we can use functions to compute the embeddings. In Sections \ref{sec:Experiments} and \ref{sec: Results}, we present the results of our approaches and compare them with baseline models.

\section{Related Work}
\label{sec: Related work}
As a reminder, several categories of KGE models can be found in the literature. This includes:
\paragraph{\textbf{Translational models}}  The foundational work in translational models was established by TransE \cite{bordes2013translating}. In TransE, given a triple $\triple{h}{r}{t} \in K$, the optimization objective involves minimizing the score $\|\mathbf{h}+\mathbf{r}-\mathbf{t}\|$, where $\mathbf{h},\mathbf{r},\mathbf{t}\in\mathbb{R}^d$. A higher score is assigned if $\triple{h}{r}{t}$ holds in $K$, and a lower score otherwise. To resolve TransE shortcomings e.g. TransE cannot model reflexive relationship \cite{chen2020knowledge,ji2015knowledge,wang2014knowledge}. Some variants have been proposed with similar ideas but different projection strategies. TransH \cite{wang2014knowledge} embeds knowledge graphs by projecting entities and relations onto hyperplanes specific to each relation. TransR \cite{lin2015learning} introduces separate spaces for entities and relations, connected by a shared projection matrix. Meanwhile, TransD \cite{ji2015knowledge} employs independent projection vectors for each entity and relation, reducing computational complexity compared to TransR. RotatE \cite{sun2019rotate} embeds entities and relations into complex space and replaces the addition operation in TransE with complex multiplication. TransG \cite{xiao2016knowledge} enhances this idea by incorporating probabilistic principles, integrating Bayesian nonparametric Gaussian mixture models and Gaussian distribution covariance. Meanwhile, TranSparse \cite{ji2016knowledge} introduces adaptive sparsity to transfer matrices, aiming to address heterogeneity issues within KGs.

\paragraph{\textbf{Rotational models}} In contrast to translational models, rotational models \cite{cao2021dual}, use the power of bilinear transformation for capturing complex interactions and correlations between entities and relations in knowledge graphs. This category of model was inaugurated by RESCAL \cite{nickel2011three} which models entities and relations using matrices, representing relationships as bilinear interactions between entity vectors. DistMult \cite{yang2014embedding} simplifies RESCAL by Representing entities and relations as low-dimensional vectors instead of matrices, employing the dot product to compute scores for triples. Although DistMult is very accurate in handling symmetric relations, it performs poorly for anti-symmetric relations. ComplEx \cite{trouillon2016complex} tackles this drawback and uses complex embeddings to model asymmetric relations, capturing both the interactions and correlations between entities and relations. 
Similarly to ComplEx, RotatE \cite{sun2019rotate} embeds entities and relations into complex space, representing relationships as rotations from head to tail entities. A variation of RotatE is RotE \cite{chami2020low} which focuses on learning rotation operations to capture relational patterns. 
QuatE \cite{zhang2019quaternion}, extends RotatE, DistMult and ComplEx by using quaternion embeddings, enabling more expressive representations of relationships in hypercomplex space. OctE Further extends QuatE by employing octonion embeddings, offering even richer and more complex representations of relationships. However, the scaling effect in the octonion and quaternion space can be a bottleneck for QuatE and OctE hence, QMult and OMult \cite{demir2021convolutional} solve this issue thanks to the batch normalization technique.  Dual quaternion methods like DualE \cite{cao2021dual} use dual quaternions \cite{jia2013dual} to embed entities and relations, offering a representation that combines the advantages of translation and rotation.

\paragraph{\textbf{Hyperbolic models}} The literature also features hyperbolic embedding methods that leverage hyperbolic geometry's properties to capture hierarchical structures in KGs. Among them, 
RotH \cite{chami2020low} extends RotatE by representing entities and relations in hyperbolic space.
 MuRP \cite{balazevic2019multi}, which maps entities and relations from a knowledge graph onto a hyperbolic space. MuRE \cite{balazevic2019multi}, which is a variant of MuRP, operates in Euclidean space, offering a simpler alternative for certain types of knowledge graphs.

\paragraph{\textbf{Deep learning models}} This group of models uses convolutional neural networks' power to encode entity and relation information simultaneously. Among them we have ConvE \cite{dettmers2018convolutional} which applies 2D convolutional filters to capture local patterns in the entity-relation space, followed by max-pooling to extract global features. Similar to ConvE, ConvO and ConvQ \cite{demir2021convolutional} also employ convolutional neural networks, but they are built upon OMult and QMult respectively.

Each of these models offers a unique approach to learning representations of entities and relations in knowledge graphs. For a more detailed review of KGE models, we recommend recent surveys \cite{DBLP:journals/tnn/JiPCMY22, DBLP:journals/tkde/WangMWG17}. However, existing methods typically represent entities and relations as static vectors, focusing mainly on operations such as multiplication, addition, scaling, and batch normalization with these vectors. 

% While such representations are practical, they are inherently static. \sh{why this is a problem?}
\paragraph{Notations:}
In the following parts of the paper, unless stated elsewhere,

\begin{itemize}
\item We denote entities and relations with minuscule letters, e.g.: $h$. Embeddings are denoted using a bold font, e.g.: the embedding of $h$ is denoted $\mathbf{h}$.
    % \item $f^n(x) = (f\circ f \underset{(n-2)}{\underbrace{\circ \cdots \circ}} f)(x)$, $\forall n\in\mathbb{N}$
   \item $\fspace(X,Y)$ the set of all function defined from $X$ to $Y$.
   \item $\Omega$ is a bounded and closed domain of $\mathbb{R}$ and $\| \Omega\|$ represent it length. e.g.: $\| [a,b] \| = b-a$.
   \item $\fspace(X)$ is the set of all function defined from $X$ to $X$.
   \item $e^{ix} = \cos x+  i\sin x$ with $i^2 = -1$.
   \item $\mathcal{R}e$ is the real part of a complex number i.e. $\mathcal{R}e(a+ib)=a$.
   \item For any vector $u$ and $v\in\mathbb{R}^n$, the product $uv$ represent the element wise multiplicatioon between $u$ and $v$.
\end{itemize}

\section{Function Space}
\label{sec:Function space}

A function  \cite{gelfand1990functions, hesselgreaves1970functions, kufner1977function} can be defined as a mathematical relationship or correspondence between two sets of elements. In other words, a function maps each input value to a unique output value. Formally, a function $f$ from a set $X$ (the domain) to a set $Y$ (the range) is denoted as $f:X\to Y$. It associates each element $x$ in the domain $X$ with one element $y$ in the range $Y$, written as $y=f(x)$. 

A function space \cite{kufner1977function}  denoted \fspace  is a space that consists of functions as its elements. In other words, it is a set of functions with certain properties defined on a given domain. That is, $\fspace(X,Y) = \{f, \  f:X\to Y\}$. They can vary widely depending on the specific properties and structures imposed on the functions within them. In this work we focus on function space that consists of $p$-integrable functions denoted $\mathcal{L}^p$ \cite{carmeli2006vector}.

\subsection{Integrable Functions}

A function $f$ is said to be $p$-integrable ($p\in\mathbb{N}$) over a domain $\Omega$ and we note $f\in\mathcal{L}^p(\Omega)$ iff the function $f$ to the power $p$ is Lebesgue integrable \cite{kufner1977function} that is,
\begin{align}
   f\in\mathcal{L}^p(\Omega)\iff \int_{\Omega} \| f(x)\|^p dx < +\infty.
\end{align}

For any $1\leq p< \infty$, $\mathcal{L}^p$ spaces are also Banach space. That is, they are all vector space with an equipped norm $\|\cdot\|_p$ defined as:  \begin{align}
        \forall f\in\mathcal{L}^p(\Omega), \|f\|_p = \Bigg(\int_{\Omega} \| f(x)\|^p dx\Bigg)^{\frac{1}{p}}.
    \end{align}

However, it's noteworthy that they do not universally qualify as Hilbert spaces, except in the special case where $p=2$, as highlighted by Young \cite{young1988introduction}. In such case they are equipped with a scalar product $\langle\cdot,\cdot\rangle_{\mathcal{L}^2(\Omega)}$ defined as:
\begin{align}
    \forall f, g\in\mathcal{L}^2(\Omega), \langle f,g\rangle_{\mathcal{L}^2(\Omega)} = \int_{\Omega} f(x)g(x)dx.
\end{align}
A straight example of integrable functions is polynomial functions.

\subsection{Polynomial Functions}
A polynomial function represents a distinctive class of mathematical functions expressed as the sum of individual terms.
We denote the space $\mathbb{R}_{deg}[x]$ the space of all polynomial functions of degree at most $deg\in\mathbb{N}$ with coefficients in $\mathbb{R}$. That is, \begin{align*} \forall P\in\fspace(\mathbb{R}), 
    P\in\mathbb{R}_{deg}[x] \iff \exists (a_i)_{i=0}^{deg}\in\mathbb{R}, P(x) = \sum_{i=0}^{deg}a_ix^i.
\end{align*}

 Clearly, for any bounded domain $\Omega\subseteq\mathbb{R}$ , $\mathbb{R}_{deg}[x]\subseteq \mathcal{L}^2(\Omega)$, $ \forall \ deg\in\mathbb{N}$. Therefore, given $P_1, P_2\in \mathbb{R}_{deg}[x]$ such that $P_1(x) = \sum_{i=0}^{deg}a_ix^i$, and $P_2(x) = \sum_{i=0}^{deg}b_ix^i$ their scalar product can be computed as:
\begin{align}
    \langle P_1(x), P_2(x)\rangle_{\mathcal{L}^2(\Omega)} &=\int_{\Omega}P_1(x)P_2(x)dx\\
    % & = \int_{\Omega} \sum_{i=0}^{deg}a_ix^i \times \sum_{j=0}^{deg}b_jx^j dx\\
    % & = \int_{m}^{M} \sum_{i,j=0}^{deg}a_ib_jx^{i+j} dx\\
    % & = \sum_{i,j=0}^{deg}a_ib_j\int_{m}^{M}x^{i+j} dx\\
    & = \sum_{i,j=0}^{deg} \dfrac{a_ib_j (M^{i+j}-m^{i+j})}{1+i+j}. \label{EqM:29}
    \end{align}
Here, $M$ and $m$ are respectively the upper and lower bounds of $\Omega$. See  the appendix in Section \ref{app:Appendix} to see how this is derived.
We can then derive the norm of all polynomial functions as:

\begin{align}
    \| P_1(x)\|^2_{\mathcal{L}^2[0,1]}& =   \langle P_1(x), P_1(x)\rangle_{\mathcal{L}^2[0,1]}\\
    & = \sum_{i=0}^{deg} \dfrac{a_i^2}{1+2i} + \sum_{\underset{i\neq j}{i,j=0}}^{deg} \dfrac{a_ia_j}{1+i+j}. \label{EqM:28}
\end{align}

These properties of polynomial functions are interesting, as their scalar product and norm can be computed without approximations, which is not always possible in $\mathcal{L}^2$. Therefore, for other types of functions, we suggest using integral approximations.

\subsection{Integral Approximation}

\begin{table}[tb]
\caption{Points and weights for adaptive Gaussian quadrature}
\centering
\resizebox{1.\linewidth}{!}{\begin{tabular}{cccccc}
\toprule
$k$  &   1   &   2   & 3 &   4   & 5 \\ \midrule
Points ($x_k$)   &   $ -\dfrac{1}{3}\sqrt{5 - 2\sqrt{\dfrac{10}{7}}}$     & $ -\dfrac{1}{3}\sqrt{5 + 2\sqrt{\dfrac{10}{7}}}$  &   $0$ &  $\dfrac{1}{3}\sqrt{5 - 2\sqrt{\dfrac{10}{7}}}$    & $\dfrac{1}{3}\sqrt{5 + 2\sqrt{\dfrac{10}{7}}}$  \\  \midrule
Weights ($w_k$) & $\dfrac{322 + 13\sqrt{70}}{900}$  & $\dfrac{322 - 13\sqrt{70}}{900}$  & $\dfrac{128}{225}$  & $\dfrac{322 + 13\sqrt{70}}{900}$  & $\dfrac{322 - 13\sqrt{70}}{900}$ \\ \bottomrule
\end{tabular}}

\label{tab:gaussquad}
\end{table}

Integral approximation is a numerical technique used to approximate the value of a definite integral by partitioning the interval of integration into subintervals and approximating the area under the curve using rectangles (for rectangle methods) and trapezoid (for trapezoid methods) \cite{burden1997numerical}
% \begin{itemize}
%     \item Rectangle 
%     \begin{align}
%         \int_{a}^{b} f(x)dx \approx \frac{b-a}{n}\sum_{k=0}^{n-1} f(a+k\frac{b-a}{n})
%     \end{align}
%     \item Trapezoid
%     \begin{align}
%         \int_{a}^{b} f(x)dx \approx \frac{b-a}{2n} \bigg( f(a) + 2\sum_{k=0}^{n-1} f(a+k\frac{b-a}{n})+f(b) \bigg)
%     \end{align}
   
% \end{itemize}

However, these approaches have the main disadvantage of not guaranteeing a maximal error. To address this limitation, we follow the methodology of works on numerical integration, by employing adaptive Gaussian quadrature \cite{golub1969calculation} with 5 nodes for the approximation of integrals. In adaptive Gaussian quadrature, an integral over the domain $[-1,1]$ can be approximated as follows: 
\begin{equation}
    \int\limits_{-1}^1 f(x)dx \approx \sum\limits_{k=1}^5 w_kf(x_k),
\end{equation}
where $w_k \in \mathbb{R}$ are weights and $x_k \in [-1, 1]$ are the nodes.
 The nodes and weights for the interval $[-1, 1]$ are provided in Table \ref{tab:gaussquad}. Using a change of variable, we can derive the Gaussian quadrature formula for an integral over any finite bounds, $a$ to $b$, as:
\begin{align}
    \int_{a}^{b}f(x)dx& \approx %\frac{b-a}{2}\int_{-1}^1f(\frac{b-a}{2}x+\frac{b+a}{2})dx\\
     \frac{b-a}{2}\sum_{k=1}^5 w_kf\bigg(\frac{b-a}{2}x_k+\frac{b+a}{2}\bigg). 
\end{align}

\section{Methodology}
\label{sec:Methodology}

\begin{table}[tb]
\centering
\caption{Embedding spaces and scoring function of  some state-of-the-art KGE models}
\label{tab:comp_time}
\begin{tabular}{lcc}
\toprule
\textbf{Models}   & \textbf{Embedding Space}  &\textbf{ Scoring Function} \\ \midrule

\textbf{\approach{}$_{n}$}  &  $\mathbb{R}_n[x]$ & $\langle \mathbf{h}(x)\otimes\mathbf{r}(x),\mathbf{t}(x)\rangle_{\mathcal{L}^2(\Omega)}$ \\

\textbf{\approach{}$^i_{n}$}  &  $\mathbb{C}_n[x]$ & $\langle\mathbf{h}(x)\otimes\mathbf{r}(x),\mathbf{t}(x)\rangle_{\mathcal{L}^2(\Omega)}$ \\

\textbf{\approach}  &  $\mathcal{L}^2(\Omega)$ & $\langle\big(\mathbf{h}\circ \mathbf{r}\big)(x),\mathbf{t}(x)\rangle_{\mathcal{L}^2(\Omega)}$ \\

\midrule

TransE & $\mathbb{R}^d$  & $\|\mathbf{h}+\mathbf{r}-\mathbf{t}\|$ \\ 

DistMult & $\mathbb{R}^d$     &  $\langle\mathbf{h}, \mathbf{r}, \mathbf{t}\rangle$                                    \\
RotatE     &    $\mathbb{R}^d$    & $ \|\mathbf{h}\circ \mathbf{r}-\mathbf{t} \|$                                    \\      
CompEx   & $\mathbb{C}^{d/2}$   & $\mathcal{R}e\langle\mathbf{h}, \mathbf{r}, \mathbf{t}\rangle$                                               \\
QMult    & $\mathbb{H}^{d/4}$   &    $\mathbf{h}\otimes \mathbf{r}^{\triangleright}\cdot \mathbf{t}$               \\ 
OMult    & $\mathbb{O}^{d/8}$  & $\mathbf{h}\otimes \mathbf{r}\cdot \mathbf{t}$               \\      
DualE    & $\mathbb{H}^{d/4}$  &     $\langle \mathbf{h}\underline{\otimes}\mathbf{r},\mathbf{t}\rangle $                  \\

HolE &  $\mathbb{R}^d$ & $\langle\mathbf{r}, \mathbf{h}\star \mathbf{t} \rangle$\\
\bottomrule
\end{tabular}
\end{table}

Let us consider the problem of embedding a KG into a  $d$- dimensional vector space, ensuring compatibility with existing methods. In Table \ref{tab:comp_time}, we show the embedding space where each of the state-of-the-art models operates. We initiate the exploration of functional embeddings by employing polynomial functions. This choice is grounded in the fundamental theorem of real analysis, known as the Weierstrass approximation theorem, which asserts that any function, no matter how complex, can be accurately approximated by polynomial functions \cite{rudin1964principles, ross2013elementary}. Given their simplicity and universal approximating power, polynomial functions serve as an ideal starting point for our investigation into functional embeddings. We then extend this idea into complex spaces, leveraging the expressive power of trigonometric functions. This serves as a natural progression before delving into more complex function representations, notably employing Neural Networks which is a kind of generalization of polynomial embeddings.

\subsection{Embedding with Polynomial Functions}
We first show how to embed KGs using polynomial functions. Specifically, when embedding in polynomial space $\mathbb{R}_n[x]$, our approach, \approach, seamlessly transforms into $\approach{}_n$, where $n$ denotes the degree of the polynomial utilized for representing the embeddings.

\begin{itemize}
\item Let $\triple{h}{r}{t}\in \mathcal{K}$
\item We compute the embedding of $\texttt{h}, \texttt{r}$ and $\texttt{t}$ in  $\mathbb{R}^m_{deg}[x]$ where $m = \frac{d}{deg+1}$ and $deg\in\mathbb{N}$ is a hyper-parameter s.t. $0\leq deg \leq d-1$ as follows:  
\begin{eqnarray*}
\mathbf{h}^{}(x) = \sum_{i=0}^{deg}a_ix^{i},
\mathbf{r}^{}(x) = \sum_{i=0}^{deg}b_ix^{i},
\mathbf{t}^{}(x) = \sum_{i=0}^{deg}c_ix^{i}
\end{eqnarray*}
where 
$a_{(\cdot)}$, $b_{(\cdot)}$ and $c_{(\cdot)} \in \mathbb{R}^m$. Hence,
\end{itemize}
\begin{align}
    \approach{}_{deg} (\triple{h}{r}{t}) & = \langle\mathbf{h}(x)\otimes \mathbf{r}(x),\mathbf{t}(x)\rangle_{\mathcal{L}^2(\Omega)}\\
     &= \int_{\Omega} \mathbf{h}(x)\mathbf{r}(x)\mathbf{t}(x)dx\\
    & = \int_{m}^M \sum_{i,j,k = 0}^{deg} a_i  b_j  c_k x^{i+j+k} dx\\
    % & =  \sum_{i,j,k = 0}^{deg} a_ib_jc_k \int_{\Omega} x^{i+j+k} dx\\
    & =  \sum_{i,j,k = 0}^{deg}\frac{ a_i b_j  c_k (M^{1+i+j+k}- m^{1+i+j+k}) }{1+i+j+k} . 
\end{align}
$M$ and $m$ represent the upper and lower bounds of $\Omega$.
\subsection{Embedding with Trigonometric Functions}

As shown with the ComplEx model \cite{trouillon2016complex}, the complex space is a prominent space to compute the embeddings as the imaginary part can be used to model anti-symmetric relations. We extend this idea by using a trigonometric function to represent the embedding. Let be $\mathbb{C}[x] = \left\{f_r + if_c, \text{where } i^2 = -1 \text{ and } f_r, f_c\in\mathcal{L}^2(\Omega) \right\}$. For any $ f\in \mathbb{C}[x]$, We have 
\begin{equation}
    ||f||^2 = \int\limits_0^1 f(x)f^*(x)dx
    = ||f_r||^2_{\mathcal{L}^2(\Omega)} + ||f_c||^2_{\mathcal{L}^2(\Omega)},
\end{equation}
where $f^* = f_r - if_c$.

For $m\in2\mathbb{N}$, we define the space $$  \mathbb{C}_n^m[x] = \left\{ \sum_{k=0}^n\alpha_k e^{ikx}, \text{ where } \alpha_k\in\mathbb{C}^{m/2} \right\}\subseteq \mathbb{C}[x]$$

When computing the embedding in complex space, \approach reduces to $\approach{}^{i}_n$. For simplicity, here we choose $\Omega = [0,1].$% where $i$ satisfies $i^2 = -1$.

\begin{itemize}
    \item Let $\triple{h}{r}{t}\in K$
    \item  We compute the embedding of $\texttt{h}, \texttt{r}$ and $\texttt{t}$ in  $\mathbb{C}^m_{deg}[x]$ where $m = \frac{d}{deg+1}$ and $deg\in\mathbb{N}$ is a hyper-parameter s.t. $0\leq deg \leq d-1$.  
    \begin{eqnarray*}
    \mathbf{h}(x) = \sum_{k=0}^{deg} a_ke^{k ix},
    \mathbf{r}(x) = \sum_{k=0}^{deg} b_ke^{k ix},
    \mathbf{t}(x) = \sum_{k=0}^{deg} c_ke^{k ix}
    \end{eqnarray*}

    where, 
$a_{(\cdot)}$, $b_{(\cdot)}$ and $c_{(\cdot)} \in \mathbb{C}^{m/2}$. Hence,
\begin{align*}
    \approach{}^{i}_n(h,r,t) &= \mathcal{R}e\bigg(\langle{\mathbf{h}(x)\otimes\mathbf{r}(x)},{\mathbf{t}(x)}\rangle_{\mathcal{L}^2(\Omega)}\bigg) \\
   %  & = \left |\int_{0}^{1}\mathbf{h}(x)\mathbf{r}(x)\mathbf{t}(x)dx\right|^2\\
   %  & =\left| \int_{0}^{1}\sum_{i,j,k=0}^{d-1}a_i^ha_j^ra_k^t\exp\Big(\delta x(i+j+k)\Big)  \right|^2\\
   %  & = \left| \sum_{i,j,k=0}^{d-1}a_i^ha_j^ra_k^t  \int_{0}^{1}\exp\Big(\delta x(i+j+k)\Big)\right|^2\\
   %  & = \left| a_0^ha_0^ra_0^t + \sum_{i,j,k=1}^{d-1}\dfrac{a_i^ha_j^ra_k^t}{i+j+k}\Big[\Big(1-\sin(i+j+k)\Big)+\delta\cos(i+j+k)\Big]  \right|^2\\
   % & = \Bigg(a_0b_0c_0 + \sum_{u,v,w=1}^{deg}\dfrac{a_ub_vc_w}{u+v+w}\Big(1-\sin(u+v+w)\Big)\Bigg)^2 \\
   % &+ \Bigg(\sum_{u,v,w=1}^{deg}\dfrac{a_ub_vc_w}{u+v+w}\cos(u+v+w)\Bigg)^2
   % & = \mathcal{R}e\bigg( a_0b_0c_0 -  \sum_{u,v,w=1}^{deg}\dfrac{a_ub_vc_w (\sin(u+v+w))}{u+v+w} \bigg)
& = \mathcal{R}e\bigg( a_0b_0c_0  + \sum_{u,v,w=1}^{deg} \frac{a_ub_vc_w}{u+v+w}\big(1-e^{i(u+v+w)}\big)\bigg)
\end{align*}

\end{itemize}
Step-by-step derivation of $\approach{}^{i}_n$ is provided in the Appendix.
Some direct consequences of these scoring formulations are the following theorems:   
\begin{theorem}
$\forall \triple{h}{r}{t}\in \KG$, 
\label{theorem: theorem1}
   \begin{align*}
        &\text{If} \ \|\Omega\| = 1, \approach{}_0(\triple{h}{r}{t}) = \text{DistMult}(\triple{h}{r}{t}) \\
        & \text{If} \ \|\Omega\| = 1, \approach{}^i_0(\triple{h}{r}{t}) =\text{ComplEx}(\triple{h}{r}{t}) 
   \end{align*}
  
\end{theorem}

\begin{theorem}
\label{theorem: theorem2}
$ \forall \triple{h}{r}{t}\in \KG\ s.t. \ \triple{t}{r}{h}\in \mathcal{K},$
     \begin{align*}
    &\approach{}_n( \triple{h}{r}{t}) = \approach{}_n( \triple{t}{r}{h})\\
     &\approach{}_n^i( \triple{h}{r}{t}) = \approach{}_n( \triple{t}{r}{h})
\end{align*}
\end{theorem}

Apart from being able to generalize DistMult and ComplEx (see Theorem \ref{theorem: theorem1}), $\approach{}_n$ and $\approach{}_n^i$ also have the advantage to model symmetric relations as highlighted in Theorem \ref{theorem: theorem2}. However, this last advantage is also their main disadvantage as it will be assigned the same score to any permutation of $\triple{h}{r}{t}$ even if this does not belong to $\mathcal{K}$ for instance, $\approach{}_{n} (\triple{h}{r}{t}) = \approach{}_{n} (\triple{r}{t}{h})\ \forall n\in\mathbb{N}$. To solve this problem, we consider using two approaches:

\begin{compactitem}
    \item Leveraging function composition in lieu of conventional function multiplication.
    \item Incorporating an activation function to amplify non-linearity within the embeddings.
\end{compactitem}

% \begin{frame}{Polynomial Embedding}{Preliminary Results}
%     \begin{center}
%         \includegraphics[scale=0.45]{images/plot_degree_poly.pdf}
%     \end{center}
% \end{frame}

\subsection{Embedding with Neural Networks}

\begin{figure*}[htb]
    \centering
    \includegraphics[width=.7\linewidth, height = 2.6cm]{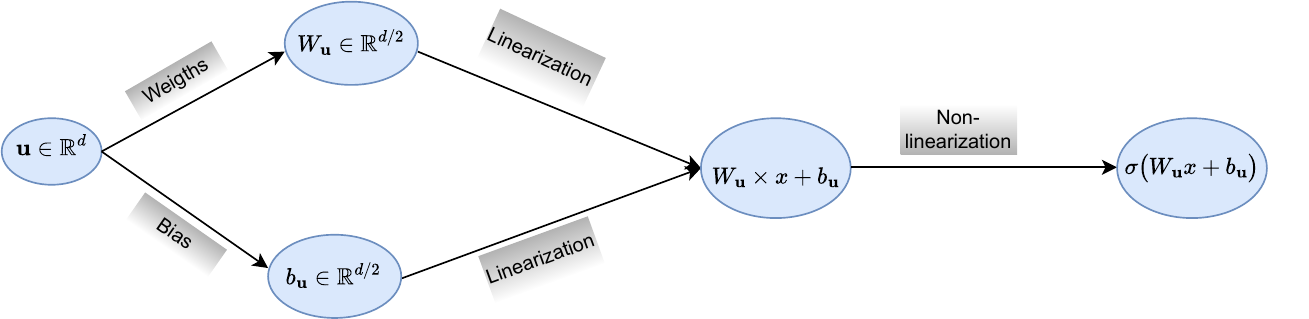}
    \caption{\approach architecture with a single layer}
    % \vspace{.5cm}
    \label{fig:FMult arch.}
\end{figure*}

Building upon the foundation laid by polynomial functions, we now delve into a more sophisticated approach by harnessing the power of Neural Networks for embedding. While polynomial functions offer simplicity and certain approximation capabilities, Neural Networks present a paradigm shift with their ability to capture highly complex and nonlinear relationships inherent in the data as mentioned by Dongare et. al. in \cite{dongare2012introduction}. Here, we explore how entities and relations in KGs can be represented as Neural Networks. Figure \ref{fig:FMult arch.} shows a summarization of \approach with a single layer.

\subsubsection{\textbf{Entities representation}}

Considering the problem of embedding into a $d$-dimensional vector space, for $\mathbf{u} = {\mathbf{h},\mathbf{r}}$ or $\mathbf{t}$, we initialize the weights and bias of the neural network as follows:
\begin{align}
    W_{\mathbf{u}} = \begin{pmatrix}
        W_{\mathbf{u_1}}\\
        W_{\mathbf{u_2}}\\
        \vdots \\
        W_{\mathbf{u_{d/2}}}\\
    \end{pmatrix} \text{ and }  b_{\mathbf{u}} = \begin{pmatrix}
        b_{\mathbf{u_1}}\\
        b_{\mathbf{u_2}}\\
        \vdots \\
        b_{\mathbf{u_{d/2}}}\\
    \end{pmatrix}
\end{align}
i.e. $W_{\mathbf{u}}, b_{\mathbf{u}} \in\mathbb{R}^{d/2}$.\\

We represent $\mathbf{u}$ as a Neural Network with $n$ layers as:

% \begin{enumerate}
%     \item If $n = d/2$ (Maximal number of layers)
%     \begin{align}
%         \textbf{u}(x) = \sigma(W_{\mathbf{u_n}}x+b_{\mathbf{u_n}})\circ\cdots\circ\sigma(W_{\mathbf{u_1}}x+b_{\mathbf{u_1}})
%     \end{align}
%     \item If $n = d/4$
    
%     \begin{align}
%     \mathbf{u}(x) =  \begin{pmatrix}
%        \sigma( W_{\mathbf{u_n}}^{}x+b_{\mathbf{u_n}}^{})\underset{(n-1)}{\underbrace{\circ \cdots \circ}} \sigma( W_{\mathbf{u_1}}^{}x+b_{\mathbf{u_1}}^{})\\
%        \sigma( W_{\mathbf{u_{d/2}}}^{}x+b_{\mathbf{u_{d/2}}}^{})\underset{(n-1)}{\underbrace{\circ \cdots \circ}} \sigma( W_{\mathbf{u_{n+1}}}^{}x+b_{\mathbf{u_{n+1}}}^{})
%     \end{pmatrix}
%     \end{align}

%     \item General form
% \end{enumerate}

% \circ \cdots \circ
\begin{align}
    \mathbf{u}(x) =  \begin{pmatrix}
       \sigma( W_{\mathbf{u_n}}^{}x+b_{\mathbf{u_n}}^{})\underset{(n-1)}{\underbrace{\circ \cdots \circ}} \sigma( W_{\mathbf{u_1}}^{}x+b_{\mathbf{u_1}}^{})\\
        \sigma( W_{\mathbf{u_{2n}}}^{}x+b_{\mathbf{u_{2n}}}^{})\underset{(n-1)}{\underbrace{\circ \cdots \circ}} \sigma( W_{\mathbf{u_{n+1}}}^{}x+b_{\mathbf{u_{n+1}}}^{})\\
        \vdots \ k \ \text{components}\\
       \sigma( W_{\mathbf{u_{kn}}}^{}x+b_{\mathbf{u_{kn}}}^{})\underset{(n-1)}{\underbrace{\circ \cdots \circ}} \sigma( W_{\mathbf{u_{(k-1)n+1}}}^{}x+b_{\mathbf{u_{(k-1)n+1}}}^{})
    \end{pmatrix}
\end{align}
where $k\in\mathbb{N}$ and such that $k \times n = \frac{d}{2}$. That is, if $n = d/2$ (Maximal number of layers)
    \begin{align}
        \textbf{u}(x) = \sigma(W_{\mathbf{u_n}}x+b_{\mathbf{u_n}})\circ\cdots\circ\sigma(W_{\mathbf{u_1}}x+b_{\mathbf{u_1}})
    \end{align}

\subsubsection{\textbf{Scoring Function Derivation}}
Given a triple $\triple{h}{r}{t}\in K$, we define the scoring function of \approach as follow:

\begin{align}
    \text{\approach}(\triple{h}{r}{t}) &= \langle\mathbf{h}\circ\mathbf{r}(x), \mathbf{t}(x)\rangle_{\mathcal{L}^2(\Omega)}\\
    & = \sum_{i=1}^{(k-1)n+1}\langle\mathbf{h}^{(i)}\circ\mathbf{r}^{(i)}(x), \mathbf{t}^{(i)}(x)\rangle_{\mathcal{L}^2(\Omega)}
    % & = \int_{\Omega} \mathbf{h}\circ \mathbf{r}(x)\cdot \mathbf{t}(x) dx\\
    % & = \int_{\Omega} \sum_{i=1}^{k}\sigma_n^n\bigg(W_{\mathbf{h_i}}^{(n)}\mathbf{r}(x)+b_{\mathbf{h_i}}^{(n)} \bigg) \times  \sigma_n^n\bigg(W_{\mathbf{t_i}}^{(n)}x+b_{\mathbf{h_i}}^{(n)} \bigg)\\
    % & = \sum_{i=1}^{k} \int_{\Omega}\sigma_n^n\bigg(W_{\mathbf{h_i}}^{(n)}\mathbf{r}(x)+b_{\mathbf{h_i}}^{(n)} \bigg) \times  \sigma_n^n\bigg(W_{\mathbf{t_i}}^{(n)}x+b_{\mathbf{h_i}}^{(n)} \bigg)
\end{align}
here, $\mathbf{h}^{(i)}(\cdot)$, $\mathbf{r}^{(i)}(\cdot)$ and $\mathbf{t}^{(i)}(\cdot)$ represent the $i$-th components of $\mathbf{h}(\cdot)$, $\mathbf{r}^{}(\cdot)$ and $\mathbf{t}^{}(\cdot)$ respectively i.e.

\begin{align}
    &\mathbf{h}^{(i)}(x) = \sigma( W_{\mathbf{h_{in}}}^{}x+b_{\mathbf{h_{in}}}^{})\underset{(n-1)}{\underbrace{\circ \cdots \circ}} \sigma( W_{\mathbf{h_{(i-1)n+1}}}^{}x+b_{\mathbf{h_{(i-1)n+1}}}^{})\\
    &\mathbf{r}^{(i)}(x) = \sigma( W_{\mathbf{r_{in}}}^{}x+b_{\mathbf{r_{in}}}^{})\underset{(n-1)}{\underbrace{\circ \cdots \circ}} \sigma( W_{\mathbf{r_{(i-1)n+1}}}^{}x+b_{\mathbf{r_{(i-1)n+1}}}^{})\\
    & \mathbf{t}^{(i)}(x) = \sigma( W_{\mathbf{t_{in}}}^{}x+b_{\mathbf{t_{in}}}^{})\underset{(n-1)}{\underbrace{\circ \cdots \circ}} \sigma( W_{\mathbf{t_{(i-1)n+1}}}^{}x+b_{\mathbf{t_{(i-1)n+1}}}^{})
\end{align}
thus,
\begin{align*}
     \text{\approach}(\triple{h}{r}{t}) &= \sum_{i=1}^{(k-1)n+1}\int_{\Omega}\mathbf{h}^{(i)}\big(\mathbf{r}^{(i)}(x)  \big)\times \mathbf{t}^{(i)}(x) dx\\
     & \approx  \sum_{j=1}^5\sum_{i=1}^{(k-1)n+1} w_j\mathbf{h}^{(i)}\big(\mathbf{r}^{(i)}(x_j)  \big)\times \mathbf{t}^{(i)}(x_j) dx,
\end{align*}
where $w_j$ and $x_j$ are the weights and points in Table \ref{tab:gaussquad}.

\section{Experiments}
\label{sec:Experiments}

\subsection{Datasets}

In our experiments, we consider our approaches for link prediction tasks, thereby providing a compelling comparison with state-of-the-art models. Prior to performing this task, we first performed an exhaustive grid search to fine-tune \approach’s and $\approach{}_n$'s parameters, seeking the optimal configuration. This phase was the most challenging aspect of our research, and the results presented on large datasets do not reflect the optimal parameters due to the substantial time investment required for this process.

We perform the evaluations considering benchmark datasets KINSHIP, COUNTRIES, UMLS, NELL-995-h100, NELL-995-h50 and NELL-995-h75. The KINSHIP dataset is a KG data that focuses on familial relationships \cite{fang2013kinship}. It typically contains information about familial connections between individuals, such as parent-child relationships, sibling relationships, grandparent-grandchild relationships, and so on. The UMLS dataset is a KG data which focuses on bio-medicine \cite{trouillon2017knowledge}. Each entity is a medical concept, and the edges represent semantic relationships between these concepts, such as "is-a," "part-of," "treats," etc. The COUNTRIES dataset typically refers to a collection of data related to countries around the world \cite{fernandez2015capital}. Finally, the NELL-995-h100, NELL-995-h50 and NELL-995-h75 datasets are subsets of the Never-Ending Language Learning (NELL) dataset designed for multi-hop reasoning \cite{lin2018multi}. The statistics related to each dataset can be found in Table \ref{tab: data_set_stats}.

\begin{table}[htb]
    \centering
    \caption{Overview of benchmark datasets}
    \small
     \resizebox{\linewidth}{!}{
    \begin{tabular}{cccccc}
    \toprule
    \textbf{Dataset} & \multicolumn{1}{c}{$|\mathcal{E}|$}&  $|\mathcal{R}|$ & $|\mathcal{G}^{\text{Train}}|$ & $|\mathcal{G}^{\text{Validation}}|$ &  $|\mathcal{G}^{\text{Test}}|$\\
    % \hline
    \midrule
    WN18-RR        &40,943      &22  &86,835       &3,034&3,134\\
    FB15k-237           & 14,541     &237 & 272,115       & 17,535&20,466\\ 
    NELL-995-h50        & 34,667      &86  & 72,767       & 5,440& 5,393\\
    NELL-995-h75           & 28,085    & 114  &59,135     &4,441&4,389\\ 
    NELL-995-h100       &22,411      & 86  & 50,314      &3,763&3,746\\
    UMLS           &135      &46  &5,216       &652&661\\ 
    KINSHIP        &104      &25  &8,544       &1,068&1,074\\
    \bottomrule
    \end{tabular}}
     
     \label{tab: data_set_stats}
\end{table}

\subsection{Training Strategy and Experimental Setup}

Throughout the experiments, all models are trained on each KG data to minimize the binary cross-entropy loss function. The link prediction performance of each KGE model is evaluated using the filtered MRR (Mean Reciprocal rank), and Hits at n (H@n) with n = 1, 3 10. To ensure a fair comparison, each model is trained using a negative sampling scoring technique, maintaining a consistent ratio of 50\%. For benchmarking purposes, we adopt embedding dimensions $d$ chosen from the set $\{16,32,64,100,128\}$, as commonly used in related literature. Optimization is performed using the Adam optimizer with a learning rate of $0.02$, and each model is trained for $500$ epochs with a batch size of $1024$.

For our approaches, the uniformity in model complexity is maintained by employing the $L_2$ regularization technique \cite{cortes2012l2}. The degree $n$ of $\approach{}_n$ is chosen between $\{0, 1, 3,7\}$ and for \approach, we found that using two layers helps the model to generalize better on some datasets (KINSHIP, NELL-995-h75) however, due to limited time complexity, we considered using only a single layer with a $\tanh$ activation function to represent all entities and relations see Figure \ref{fig:FMult arch.}. %The domain where each function operates through the experiments has been set to $\Omega = [-1,1]$.

\section{Results and Discussion}
\label{sec: Results}

\begin{table*}[h!tb]
    \centering
    \caption{Link prediction results on  UMLS, KINSHIP and NELL-995-h100. Results are taken from the corresponding paper. Bold and underlined results indicate the best and second-best results respectively. The dash (-) denotes values missing in the paper}
    \small
    % \scriptsize
%\footnotesize
% \setlength{\tabcolsep}{4pt}
%\resizebox{0.9999\linewidth}{!}{
\begin{tabular}{@{}l cccc ccccc ccccc cccc c@{}}
    \toprule
     \textbf{Models} &\multicolumn{4}{c}{\textbf{UMLS}} && \multicolumn{4}{c}{\textbf{KINSHIP}}&& \multicolumn{4}{c}{\textbf{NELL-995-h100}}\\
     \cmidrule(l){2-5} \cmidrule(l){7-10} \cmidrule(l){12-15} 
        &MRR  &H@1   &H@3   &H@10 && MRR & H@1  & H@3  & H@10  && MRR & H@1  & H@3  & H@10\\
\midrule

TransE & .72 & .55 & .87& .96 && .14 & .05& .13& .30 && .51& .37 & .59 & .77

                                                           \\

TransR &.89 &.79 &.98 &\underline{.99} && .67& .53&.77 &.96 && .42& .31 & .48 & .62

                                                        \\

DistMult  &.81 & .74& .86 & .95 && .67 & .54& .75 & .93 && .77 & .66 & .86 & .96 

                                                   \\

ComplEx  & .96 & .92&.99 &\textbf{1.0} && \underline{.89}& \underline{.82}& \underline{.96}& \textbf{.99}&& \underline{.83} & .74 &.91  & \textbf{.98}
                                            \\

QMult  \cite{demir2021convolutional}  & \underline{.96} & \underline{.93} & \underline{.98}& \textbf{1.0} && .88 & .81& .94& \textbf{.99 }&& \underline{.83}& .73 & .91 & \textbf{.98}
                                                               \\

OMult \cite{demir2021convolutional}  & .95 & .91 &  .98& 1.0 && .87 & .80 & .94& \textbf{.99} && .75& .64 & .82 & .93                                      \\

MuRE& \textbf{.97 }& \textbf{.95} & \textbf{.99} & \textbf{1.0} && .82 & .72 & .89 & \underline{.98} && \underline{ .83} & .74& .90& \textbf{.98}                                  \\

MuRP& .89 & .80 & .97 & \underline{.99} &&  .75 & .62 & .84 &. 96 && \textbf{.85} & \underline{.75} &\underline{.90} &\textbf{.98}                                                    \\

HolE& .95 & .91& .99 & \textbf{1.0} && .89 & .82 & .96 &\textbf{.99}  && .71 & .59 &.79&.92                                                        \\

RotatE & .87 & .77 & .95 & .98 && .76 & .65 & .84 & .96 && .46 & .38 & .49 & .63                                       \\

TuckER& .87 &.78 &.96 &.98 && .58 & .42 & .68 & .92 && .33 & .23 & .37 & .54                                               \\

{DualE} 
&      .94 &.89 & .99& \textbf{1.0}&& .82 & .71& .91&\textbf{.99} & & .77&  .67 & .85 &.96                  \\

{ConvO} \cite{demir2021convolutional} & .90 & .82 & .98 & \textbf{1.0} && .86 & .77 & .93 & \underline{.98} && -&-&-&- \\

{ConvQ}  \cite{demir2021convolutional}  & .92 & .86 & .98 &\textbf{1.0}&& .86 & .77& .93&  \underline{.98} && -&-&-&- 
                                  \\

\midrule

\textbf{${\approach}_n$} & .75 & .65& .82& .93 && .59 & .44 & .68 & .91 && \underline{.83} & \textbf{.76} & .88 & .96                                        \\ 

\textbf{{\approach}}&  \textbf{.97 }& \textbf{.95} & \textbf{.99} & \textbf{1.0} && \textbf{.90} & \textbf{.83} & \textbf{.98} &\textbf{.99} && \textbf{.85} & \textbf{.76} & \textbf{.93} & \underline{.97}  
                                                            \\

    \bottomrule
   
    \end{tabular}%}
    %\label{table:wn_fb_yago}
    \label{tab:link_pred2} 
\end{table*} 
% \sh{a bit more here}
% In this part we mainly present the results of $\approach$ and $\approach{}_n$ for link prediction
\subsection{Embedding Dynamics}

% \begin{figure}
% \centering
% \begin{subfigure}{0.6\textwidth}
%     \includegraphics[width=.4\textwidth]{samples/Images/distmult_enities_2d.pdf}
%     \caption{DistMult}
% \end{subfigure}
% \begin{subfigure}{0.6\textwidth}
%      \includegraphics[width=.4\textwidth]{samples/Images/PolyMult_enities_2d.pdf}
%      \caption{Firts subfigure.}
% \end{subfigure}

% \begin{subfigure}{0.6\textwidth}
%     \includegraphics[width=.4\textwidth]{samples/Images/LFMult_enities_2d.pdf}
%     \caption{\approach}
% \end{subfigure}
% \begin{subfigure}{0.6\textwidth}
%     \includegraphics[width=.4\textwidth]{samples/Images/LFMult1_enities_2d.pdf}
%     \caption{$\approach{}$}
% \end{subfigure}

%  \caption{Entities representation for the COUNTRIES dataset in 2-dimension}
% \label{fig:ploted_entities}
% \end{figure}

\begin{figure}[htb] 
	
	\centering
	\subcaptionbox{DistMult}{\includegraphics[width=0.47\linewidth]{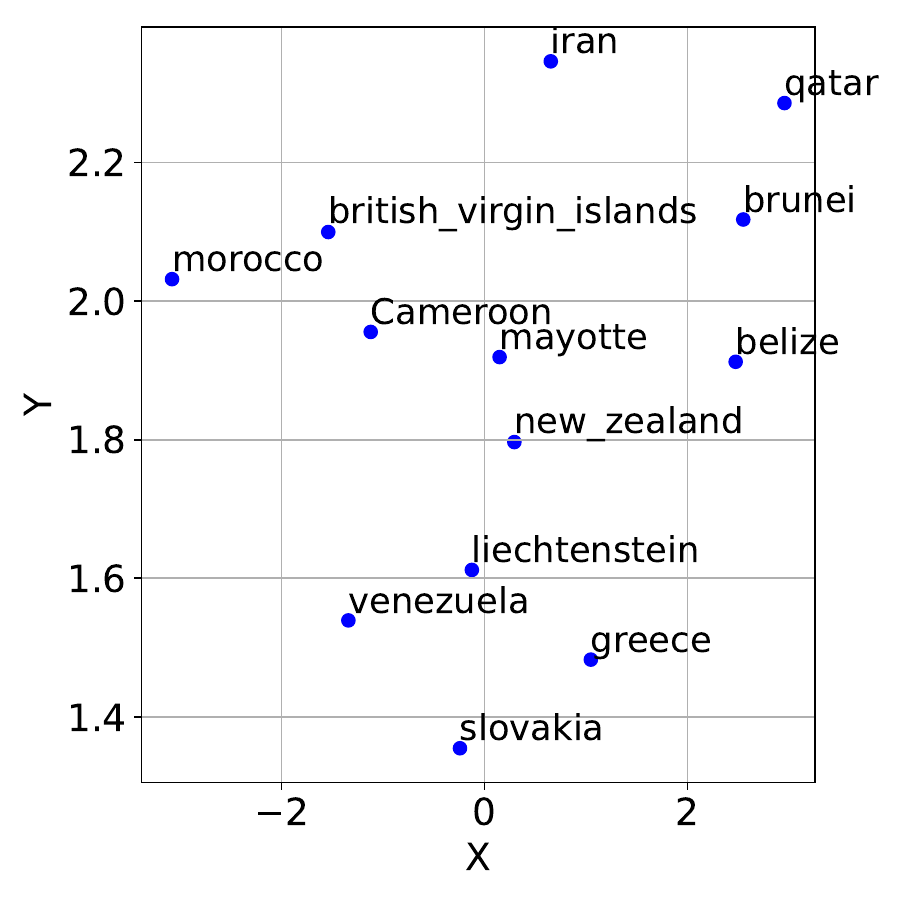}}  
	\subcaptionbox{\approach}{\includegraphics[width=0.47\linewidth]{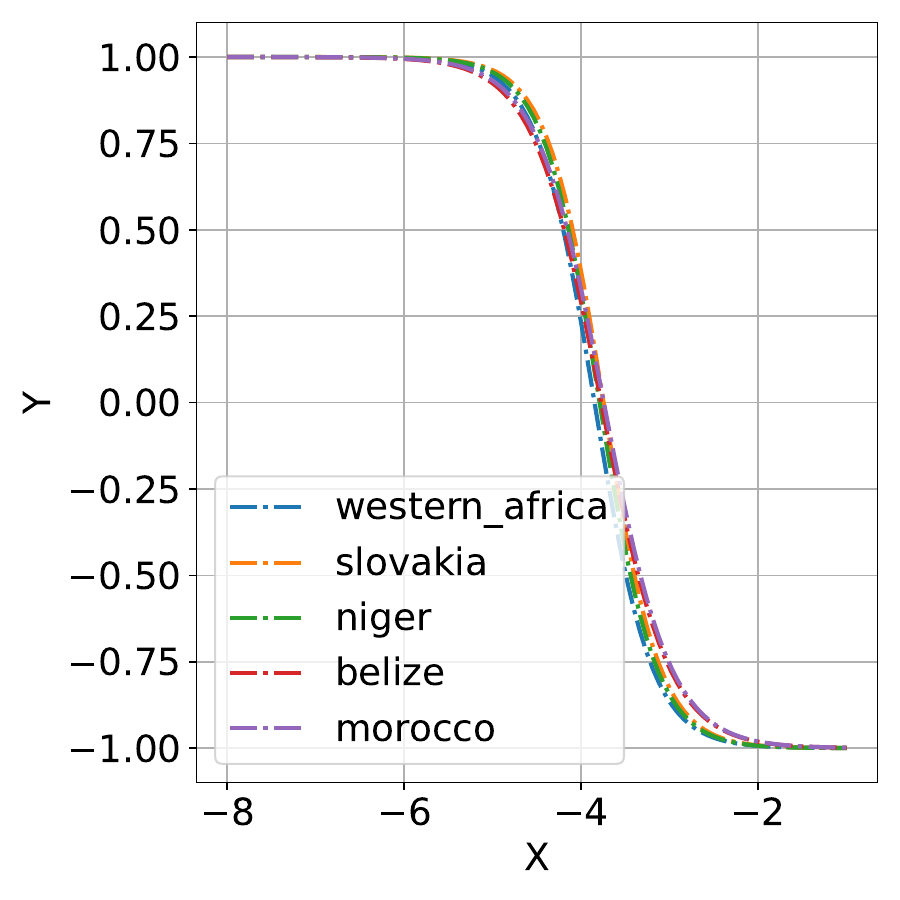}}

	\subcaptionbox{$\approach{}^i_1$}{\includegraphics[width=0.47\linewidth]{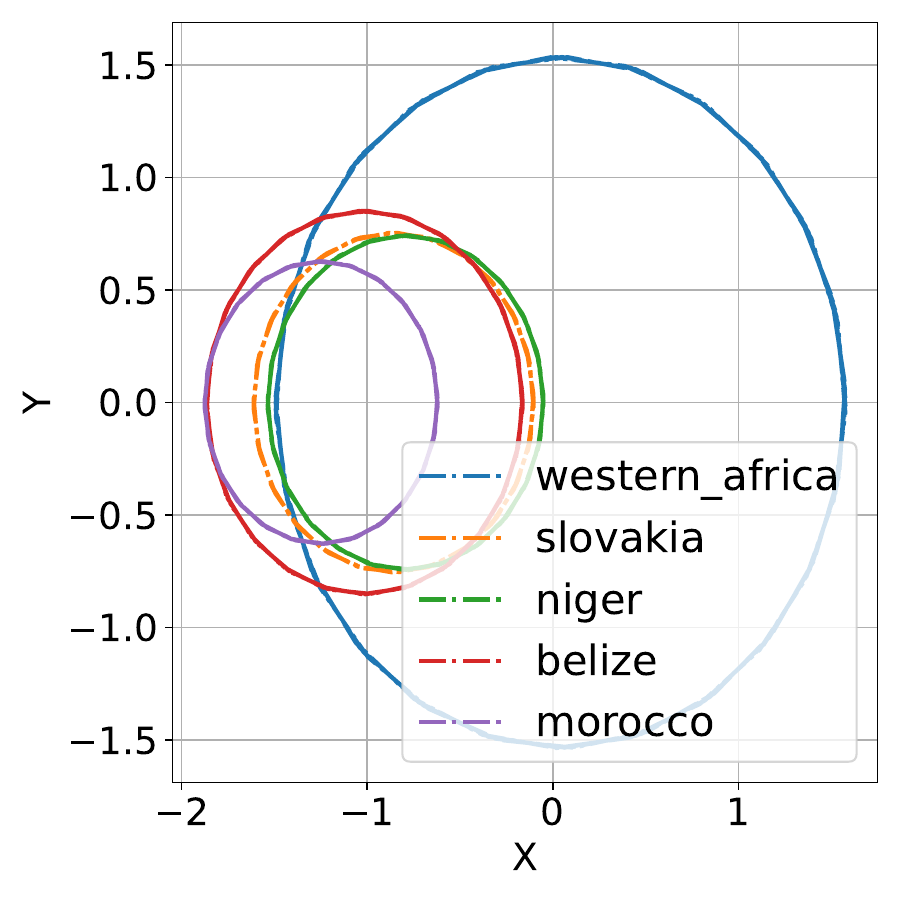}}
	\subcaptionbox{$\approach{}_2$}{\includegraphics[width=0.47\linewidth]{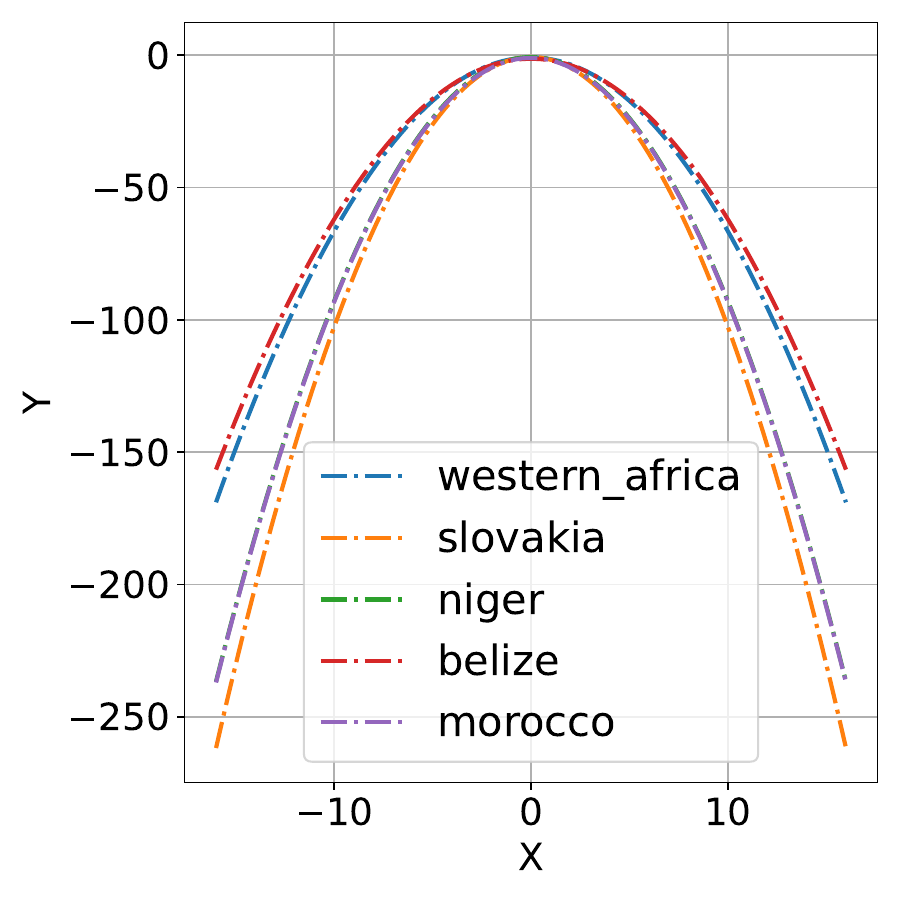}}
	
	\caption{Entities representation for the COUNTRIES dataset in 2-dimension}
	\label{fig:ploted_entities}
\end{figure}

\begin{figure}[htb]
	\centering
	\includegraphics[width=0.94\linewidth]{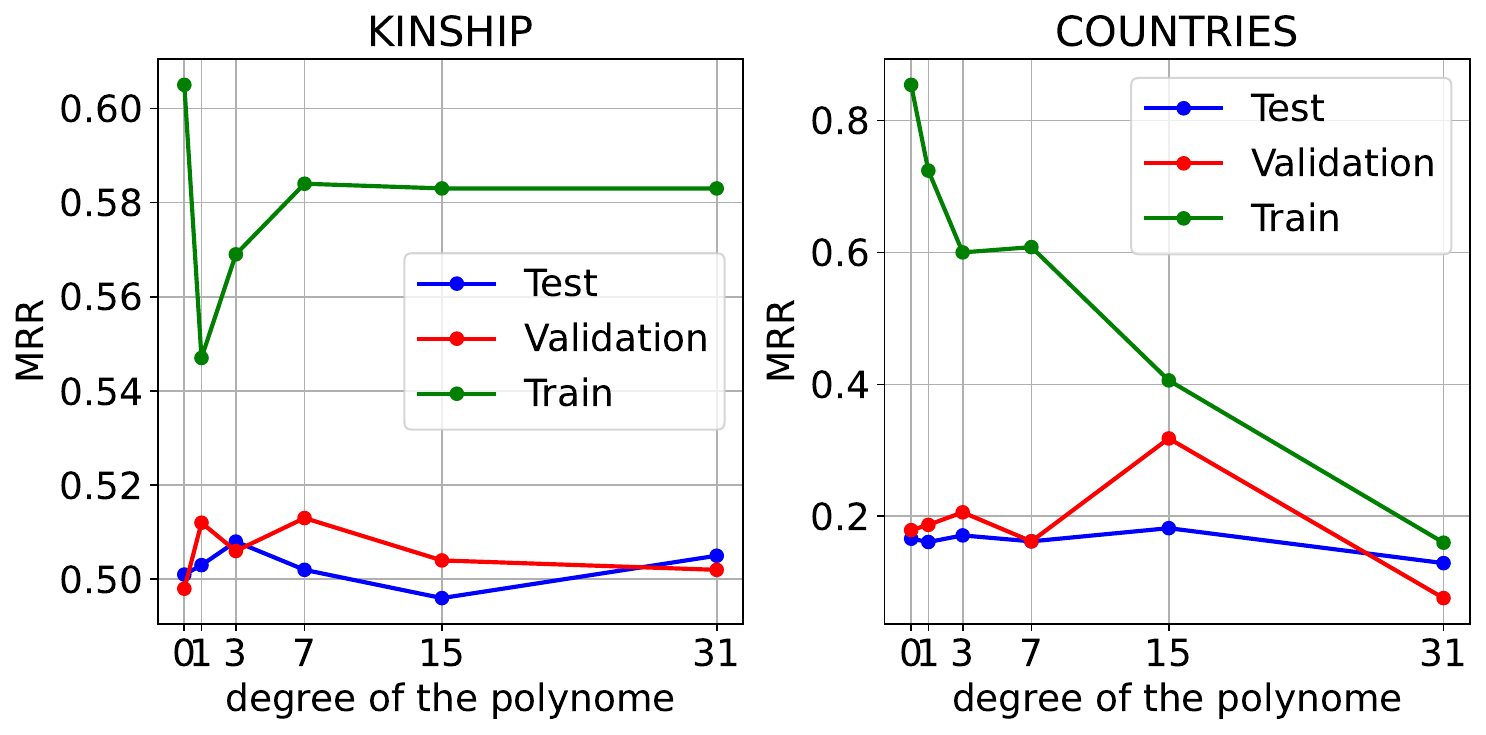}
	\caption{Impact of degrees on polynomial embeddings}
	% \vspace{.5cm}
	\label{fig:degree_polynomial}
\end{figure}

In Figure \ref{fig:ploted_entities}, we show the embeddings of the COUNTRIES dataset in two dimensions, showcasing the representation of each entity across different variants of the \approach method and the DistMult model. As expected, DistMult displays the embeddings as 2-dimensional vectors. When applying a sigmoid activation function to \approach, it effectively uses this function to represent the computed embeddings. Conversely, $\approach{}_1^i$ uses trigonometric functions, resulting in circular embeddings with varying centres and diameters determined by the trained embeddings. Further, $\approach{}_2$ employs squared functions to represent the embeddings, resulting in shapes characterized by a parabola. Notably, all embeddings share a common focus and vertex, highlighting a consistent feature across the diverse visualization techniques.

% \sh{the reference to figures needs to be somewhat clear, what is the message here?}
We also investigate in Figure \ref{fig:degree_polynomial} how the degree of $\approach_n$ impacts the embedding on  the KINSHIP and COUNTRIES datasets. More precisely, we plot the MRR achieved by $\approach_n$ on those datasets for $n = \{1,3,7,15,31\}$ at the training, test and validation phase. On the KINSHIP data, it is clear that a 3-degree polynomial function generalizes better than other degrees. An explanation for this is due to the nature of the KINSHIP data itself which mainly consists of symmetric relationships. Meanwhile, on the COUNTRIES dataset, a 15-degree polynomial offers a better generalization compared to others.

\subsection{Link Prediction Results: Comparison with the state-of-the-arts}

\begin{table}[tb]
    \centering
    \caption{Link prediction results for WN18-RR and FB15k-237 datasets. Results are taken from corresponding papers. Bold and underlined results indicate the best and second-best results respectively.}
    \small
\resizebox{\linewidth}{!}{\begin{tabular}{l cccc cccc}
    \toprule
    \textbf{Models} & \multicolumn{4}{c}{\textbf{WN18-RR}} & \multicolumn{4}{c}{\textbf{FB15k-237}} \\
    \cmidrule(lr){2-5} \cmidrule(lr){6-9} 
    & MRR & H@1 & H@3 & H@10 & MRR & H@1 & H@3 & H@10 \\
    \midrule
TransE \cite{ruffinelli2020you} & .23 & .05 & .37 & .52 & .31 & .22 & .35 & .50 \\
% TransR & & & & & & & & \\
DistMult \cite{dettmers2018convolutional} & .43 & .39 & .44 & .49 & .24 & .16 & .26 & .42 \\
ComplEx \cite{dettmers2018convolutional} & .44 & .41 & .46 & .51 & .25 & .16 & .28 & .43 \\
QMult \cite{demir2021convolutional} & .44 & .39 & .45 & .54 & .35 & .25 & .38 & \underline{.54} \\
OMult \cite{demir2021convolutional} & .45 & .41 & .48 & .54 & \underline{.38} & .25 & .38 & .53 \\
MuRE\cite{chami2020low} & .46 & .42 & .47 & .53 & .31 & .23 & .34 & .49 \\
MuRP \cite{cao2021dual} &\underline{.48}& \underline{.55} & \underline{.49} & .44 & .34 & \underline{.52} & .37 & .25 \\
% HolE & & & & & & & & \\
RotatE \cite{cao2021dual} & \underline{.48} & .43 & \underline{.49} & \underline{.57} & .34 & .24 & .38 & .53 \\
TuckER \cite{balavzevic2019tucker} & .47 & .44 & .48 & .53 & .36 & .27 & .39 & .54 \\
DualE \cite{cao2021dual} & \textbf{.49} & \textbf{.58} & .51 & .44 & .37 & \textbf{.56} & \underline{.40} & .27 \\
ConvO \cite{demir2021convolutional} & .46 & .43 & .47 & .52 & .37 & .27 & \underline{.40} & \underline{.54} \\
ConvQ \cite{demir2021convolutional} & .46 & .42 & .47 & .53 & .34 & .25 & .38 & .53 \\
\midrule
\textbf{$\approach{}_n$} & .43 & .39& .44& .50& \textbf{. 47}  & .31  & \textbf{.56}   & \textbf{.59}   \\
\textbf{$\approach$} & .43 &.27 &\textbf{.52} &\textbf{.60}   & .03 & .01 & .02&.08 \\
\bottomrule
    \end{tabular} }

    \label{tab:link_pred_FB_WN}
\end{table}

%%%%%%%%%%%%%%%%%%%%%%%%%%%%%%%%%%%%%%%%%%%%%%%%%%%%%%%%%%%%%%%%%%%%%%%%%%%%%%%%%%%%%%%%%%%%%%%%%%%%%%%%%%%%%%%%%%%%%%%%%%%%%%%%%%%%%%%%%%%%%%%%%%%%%%%%%%%%%%%%%%%%%%%%%%%%%%%%%%%%%%%%%%%%%%%%%%%%%%%%%%%%%%%%%%%%%%%%%%%%%%%%%%%%%%%%%%%%%%%%%%%%%%%%%%%%%%%%%%%%%%%%%%%%%%%%%%%%%%%%%%%%%%%%%%%%%%%%%%%%%%%%%%%%%%%%%%%%%%%%%%%%%%%%%%%%%%%%%%%%%%%%%%%%%%%%%%%%%%%%%%%%%%%%%%%%%%%%%%%%%%%%%%%%%%%%%%%%%%%%%%%%%%%%%%%%%%%%%%%%%%%%%%%%%%%%%%%%%%%%%%%%%%%%%%%%%%%%%%%%%%

\begin{table*}[h!tb]
    \centering
    \caption{Link prediction results on NELL-995-h75, NELL-995-h50 and Countries. Results are taken from the corresponding paper Bold and underlined results indicate the best and second-best results respectively.}
    \small
    % \scriptsize
%\footnotesize
% \setlength{\tabcolsep}{4pt}
%\resizebox{0.9999\linewidth}{!}{
\begin{tabular}{@{}l cccc ccccc ccccc cccc c@{}}
    \toprule
     \textbf{Models} &\multicolumn{4}{c}{\textbf{NELL-995-h75}} && \multicolumn{4}{c}{\textbf{NELL-995-h50}}&& \multicolumn{4}{c}{\textbf{COUNTRIES}}\\
     \cmidrule(l){2-5} \cmidrule(l){7-10} \cmidrule(l){12-15} 
        &MRR  &H@1   &H@3   &H@10 && MRR & H@1  & H@3  & H@10  && MRR & H@1  & H@3  & H@10\\
\midrule

TransE & .55 & .41 & .63 & .78 && .56 & .43 & .63 & .80 && .70 & .41 & .99 & 1.0

                                                           \\

TransR & .44 & .32 & .50 & .63 && .44 & .34 & .49 & .62 && .68 &  .42 & .93 & .99

                                                        \\

DistMult& .77 & .67 & .84 & .95 &&  .75 & .65 & .82 & .95 && \underline{.98} & . 97& .99 & .99

                                                    \\

ComplEx& .83 & .75 & .89 &  \underline{.97} && .79 & .70 & .87 &\textbf{.97} &&  \underline{.98} & \underline{.97} & .99 & 1.0
                                                       \\

QMult \cite{demir2023clifford} & \textbf{.94 }& \textbf{.91} & \textbf{.97} & \textbf{.98} && .67 & .58 & .73 & .82 && \underline{.98} & .96 & .99 & .99
                                                    \\

OMult\cite{demir2023clifford} & .66 & .57 & .74 & .83 &&  .67 & .58 & .73 & .82  && \underline{.98} & \underline{.97} & .99 & 1.0                                \\

MuRE &.79 & .69& .86&.96 && .78& .68 &.85 &.95 && \textbf{.99}& \textbf{.98} & 1.0 & 1.0                                                          \\
% MuRP & & & & & & & & & & & .89&.80 &.97 &1.0                                                          \\

HolE& .71& .60& .78 & .91    &&\underline{.81} & \underline{.72}& \textbf{.89}& \textbf{.97}&& \underline{.98}& .96 & .99 & 1.0                                                           \\

RotatE & .46 & .39 & .49 & .63 && .46 &.37 & .49& .63&& .77 & .61 & .93 & .99                                                   \\

TuckER& .34 & .23&.37&.54   &&.33 &.22 &.36 &.53&& .09 & .03 & .09 & .21                                                        \\

{DualE} &.77 &.68 &.85 &.94 && .75 & .63 & .83 & .95 && .95 & .91 & .98 &.99
                                                       \\

\midrule

\textbf{${\approach}_n$} & .77 & .68& .83& .95 && \textbf{.82} & \textbf{.74}& \underline{.88}& \underline{.96}& & \underline{.98} & \textbf{.98} & .99& .99                                          \\ 

\textbf{{\approach}} & \underline{.85} & \underline{.77} & \underline{.93} &\textbf{ .98} && .77 & .64 & .87 & \textbf{.97} && .93 & .86 & .99 & 1.0
                                                            \\

    \bottomrule
   
    \end{tabular}%}
    
    \label{tab:link_pred1} 
\end{table*} 

\balance

Table \ref{tab:link_pred2} presents the results on the UMLS, KINSHIP, and NELL-995-h100 datasets. On the UMLS dataset, \approach excels, achieving the highest scores across all metrics, tying with MuRE and outperforming other state-of-the-art models. $\approach{}_n $ does not have high performance, however, remains competitive. Since the UMLS data contains hierarchical relationships that are often complex, we can conclude that representing entities as neural networks is highly effective in capturing complex and non-linear relationships, which is likely why \approach achieved outstanding results across all metrics. However, polynomial functions may not be as flexible as neural networks in capturing the nuances of hierarchical and medical data. This might explain why $\approach{}_n$ performed lower than \approach and other state-of-the-art models.  On the KINSHIP data \approach leads in MRR, H@1, and H@3, outperforming all other models, while $\approach{}_n $ lags behind most state-of-the-art models. Note that the KINSHIP dataset is known for its numerous symmetric relationships, such as parent-child and sibling relationships. The result suggests that neural networks can seamlessly handle symmetry, which might explain why \approach  leads in most metrics. However, polynomial functions may not be particularly adept at capturing symmetric relationships, which could explain the middling performance of $\approach{}_n$ on this dataset leading to lower scores compared to neural networks. A similar observation can be found on the NELL-995-h100 where \approach demonstrates outstanding performance, leading in MRR, H@1, and H@3, and achieving near-perfect scores in H@10. $\approach{}_n $ also performs well on this dataset, closely following the top models.

In Table \ref{tab:link_pred_FB_WN}, we describe the results of the WN18RR and FB15k-237 datasets. On the WN18RR dataset, $\approach{}_n$ is trained with $n=0$ and performed as expected similarly to DistMult but does not outperform the state-of-the-art models. However,  $\approach$  shows a stronger performance in H@3 and H@10, indicating it is particularly effective in retrieving the correct tail entity within the top 3 and top 10 predictions of the dataset. This implies that while it may struggle to predict the exact head entity (lower H@1), it captures useful information that positions the correct entity closer to the top in ranked lists. DualE and RotatE outperform both $\approach$ and $\approach{}_n$ in terms of MRR and H@1, which highlights the efficiency of models that capture rotational and dual embeddings for this dataset. On the FB15k-237, $\approach{}_n$ is trained  using one-degree polynomials ($n=1$) and this significantly outperforms other models. Meanwhile, we observe a very poor performance of $\approach$. 
% \sh{The next sentence is not clear}
This extreme suboptimal performance is due to inappropriate hyperparameter settings for this dataset. Given the large size of the data, experimenting with various settings is very time-consuming.

Table \ref{tab:link_pred1} presents the link prediction results on the NELL-995-h75, NELL-995-h50, and Countries datasets. On the NELL-995-h75, \approach excels with second-best performance across all metrics, closely following QMult. $\approach{}_n $  also performs competitively, especially in H@10.
On the NELL-995-h50 dataset, \approach and $\approach{}_n$ both show strong performance, but in slightly different ways: \approach is most effective at predicting H@10 showing best performance with HolE and ComplEx due to the flexibility of neural networks. While $\approach{}_n$ excels in accurately identifying the most relevant connections like H@1 and H@3 and maintaining a high overall performance (MRR), indicating the efficacy of polynomial functions in capturing a mix of relationship complexities contained in the NELL-995-h50 dataset. This analysis highlights the complementary strengths of the two approaches, with neural networks providing broader predictive power and polynomial functions offering precision in ranking relevance on this particular dataset. On the Countries data, apart from TuckER, all other approaches perform incredibly well which is because the data itself contains only two relations. However, both \approach and $\approach{}_n$ demonstrate strong performance, with $\approach{}_n $ achieving near-perfect scores across all metrics. Overall, both \approach and $\approach{}_n $ demonstrate competitive performance against state-of-the-art models, with \approach often achieving second-best results and $\approach{}_n $ excelling in several metrics, particularly in the NELL-995-h50 and Countries datasets.

\approach excels in datasets with complex, hierarchical (e.g. UMLS), or symmetric relationships (KINSHIP, WN18-RR), outperforming state-of-the-art models, while $\approach{}_n$ remains competitive but less effective due to its reliance on polynomial functions. Conversely, $\approach{}_n$ significantly outperforms in datasets suited to polynomial representations (e.g. NELL-995-h50) but may underperform if hyperparameters are not well-tuned. Both approaches demonstrate robust performance in datasets with fewer relationships (e.g. COUNTRIES), with $\approach{}_n$ particularly excelling in precision.

% \sh{the description of the tables should be in order}

% \sh{a discussion summarizing the result is needed}

\section{Conclusion}

\label{sec:Conclusion}

In this work, we have developed three novel embedding methods that operate in function space: $\approach{}_n$, $\approach{}^i_n$, and $\approach{}$. This marks the first time that functional representations have been employed for embedding entities and relations in knowledge graphs. Our experimental results on eight benchmark datasets demonstrate that either polynomial-based embeddings ($\approach{}_n$) or neural network-based embeddings ($\approach{}$) can significantly improve state-of-the-art results in knowledge graph completion.

The promising outcomes suggest that functional representations provide a more flexible and expressive framework for capturing the complex dynamics within knowledge graphs. While this study has focused on $\approach{}_n$ and $\approach{}$, the potential of the trigonometric function-based approach ($\approach{}^i_n$) remains unexplored and will be the subject of future work.

% \section{Acknowledgement}
\begin{acks}
This project received funding from the European Union's Horizon Europe research and innovation programme through two grants (Marie Skłodowska-Curie grant No. 101073307 and grant No. 101070305). It was also supported by the "WHALE" project (LFN 1-04), funded by the Lamarr Fellow Network programme and the Ministry of Culture and Science of North Rhine-Westphalia (MKW NRW).
\end{acks}

\bibliographystyle{plain}
\balance
\bibliography{software}

\newpage
 
% \section{Appendix}
\appendix
\label{app:Appendix}
\section{Scores Derivation}
\subsection{Scalar Product and Norm Derivation of Polynomial Functions}

Given $P_1, P_2\in \mathbb{R}_{deg}[x]$ such that $P_1(x) = \sum_{i=0}^{deg}a_ix^i$, and $P_2(x) = \sum_{i=0}^{deg}b_ix^i$ we have:
\begin{align}
    \langle P_1(x), P_2(x)\rangle_{\mathcal{L}^2(\Omega)} &=\int_{\Omega}P_1(x)P_2(x)dx\\
    & = \int_{\Omega} \sum_{i=0}^{deg}a_ix^i \times \sum_{j=0}^{deg}b_jx^j dx\\
    & = \int_{\Omega} \sum_{i,j=0}^{deg}a_ib_jx^{i+j} dx\\
    & = \sum_{i,j=0}^{deg}a_ib_j\int_{\Omega}x^{i+j} dx\\
    & = \sum_{i,j=0}^{deg} \dfrac{a_ib_j \big(\sup \Omega^{i+j}-\inf \Omega^{i+j}\big)}{1+i+j}. \label{EqM:27}
    \end{align}
Here, $\sup \Omega$ and $\inf \Omega$ are respectively the upper and lower bounds of $\Omega$.
% % We can then derive the norm of all polynomial function as:
% % \begin{align}
% %     \| P_1(x)\|^2& =  \langle P_1(x), P_1(x)\rangle_{\mathcal{L}^2(\Omega)}\\
% %     & = \sum_{i,j=0}^{deg} \dfrac{a_ia_j \big(\sup \Omega^{i+j}-\inf \Omega^{i+j}\big)}{1+i+j}\\
% %     & = \sum_{i=0}^{deg} \dfrac{a_i^2 \big(M^{2i}-m^{2i}\big)}{1+2i} + \sum_{\underset{i\neq j}{i,j=0}}^{deg} \dfrac{a_ia_j \big(M^{i+j}-m^{i+j}\big)}{1+i+j}.
% % \end{align}

% With $M = \sup \Omega$ and $m = \inf \Omega$.
\subsection{Full Scoring Function Derivation of $\approach{}^{i}_n(h,r,t) $}
\label{subsection: full_derivatio}
 
\begin{itemize}
    \item Let $\triple{h}{r}{t}\in K$
    \item  We compute the embedding of $\texttt{h}, \texttt{r}$ and $\texttt{t}$ in  $\mathbb{C}^m_{deg}[x]$ where $m = \frac{d}{deg+1}$ and $deg\in\mathbb{N}$ is a hyper-parameter s.t. $0\leq deg \leq d-1$.  
    \begin{eqnarray*}
    \mathbf{h}(x) = \sum_{k=0}^{deg} a_ke^{k ix},
    \mathbf{r}(x) = \sum_{k=0}^{deg} b_ke^{k ix},
    \mathbf{t}(x) = \sum_{k=0}^{deg} c_ke^{k ix}
    \end{eqnarray*}

    where, 
$a_{(\cdot)}$, $b_{(\cdot)}$ and $c_{(\cdot)} \in \mathbb{C}^{m/2}$. Hence,
\begin{align*}
    &\approach{}^{i}_n(h,r,t) = A\\
    &A = \mathcal{R}e\bigg(\langle{\mathbf{h}(x)\otimes\mathbf{r}(x)},{\mathbf{t}(x)}\rangle_{\mathcal{L}_2(\Omega)}\bigg) \\
 & = \mathcal{R}e\bigg( \int_{\Omega} \sum_{k=0}^{deg} a_ke^{k ix} \times \sum_{k=0}^{deg} b_ke^{k ix} \times \sum_{k=0}^{deg} b_ke^{k ix} dx\bigg)\\
 & =  \mathcal{R}e\bigg( \int_{\Omega} \sum_{u,v,w=0}^{deg} a_ub_vc_w e^{i(u+v+w)x} dx\bigg)\\
 & =  \mathcal{R}e\bigg(  \int_{\Omega} a_0b_0c_0 dx + \sum_{u,v,w=1}^{deg}\int_{\Omega} a_ub_vc_w e^{i(u+v+w)x} dx\bigg)\\
 & = \mathcal{R}e\big( a_0b_0c_0 \|\Omega\| \big)+ \mathcal{R}e\bigg(\\ &  \sum_{u,v,w=1}^{deg} \frac{a_ub_vc_w}{u+v+w}\big(e^{i(u+v+w)\times \inf\Omega}-e^{i(u+v+w)\times \sup \Omega}\big)\bigg)
\end{align*}
\end{itemize}
 
\section{Proof of Theorems}
\begin{itemize}
\item $\text{If} \ \|\Omega\| = 1, \approach{}_0(\triple{h}{r}{t}) = \text{DistMult}(\triple{h}{r}{t})$
    \begin{proof}
        
            We assume $\|\Omega\|=1$. If $deg = 0$,  the polynomials reduce to $d$-dimentional constant vectors (as $m = \frac{d}{0+1} = d$) i.e.,
            $$ \mathbf{h}(x)=a_0,\ \mathbf{r}(x)=b_0, \ \mathbf{t}(x)=c_0. $$
            
            Therefore,  
            \begin{align}
                \approach{}_0(\triple{h}{r}{t}) &= \int_{\Omega} a_0 b_0 c_0 dx\\
                & =  a_0  b_0 c_0 \ \|\Omega\| \\
                & = a_0 b_0 c_0 \\
                & = \text{DistMult}(\triple{h}{r}{t})
            \end{align}

    \end{proof}

\item   $ \text{If} \ \|\Omega\| = 1, \approach{}^i_0(\triple{h}{r}{t}) \equiv\text{ComplEx}(\triple{h}{r}{t}) $

\begin{proof}
    \item  From Subsection \ref{subsection: full_derivatio}, we can directly derive that if $deg=0$, the second term of the summation vanish. We therefore remain with $\approach{}^{i}_0(h,r,t) =  \mathcal{R}e\big( a_0b_0c_0 \|\Omega\| \big) $ and since by hypothesis, $\|\Omega\| = 1$ we get  $\approach{}^{i}_0(h,r,t) =  \mathcal{R}e\big( a_0b_0c_0 \big) $ with $a_0, b_0$ and $c_0\in\mathbb{C}^{d/2}$ which is the definition of ComplEx.
\end{proof}
     \end{itemize} 
    
% \subsection{Proof for Theorem \ref{theorem: theorem2}}

\begin{itemize}
\item $\approach{}_n(\triple{h}{r}{t}) = \approach{}_n(\triple{t}{r}{h})$

\begin{proof}

   Using the definition of $\approach{}_n$:
   \begin{align}
        \approach{}_n(\triple{h}{r}{t})& = \int_{\Omega} \mathbf{h}(x) \mathbf{r}(x) \mathbf{t}(x) \, dx
   \end{align}
   By swapping $\mathbf{h}$ and $\mathbf{t}$:
  
   \begin{align}
        \approach{}_n(\triple{t}{r}{h}) &= \int_{\Omega} \mathbf{t}(x) \mathbf{r}(x) \mathbf{h}(x) \, dx \\
        & = \int_{\Omega} \mathbf{h}(x) \mathbf{r}(x) \mathbf{t}(x) \, dx\\
        & = \approach{}_n(\triple{h}{r}{t})
   \end{align}
   
\end{proof}
\item $\approach{}_n^i(\triple{h}{r}{t}) = \approach{}_n(\triple{t}{r}{h})$

\begin{proof}
    
   Using the definition of $\approach{}_n^i$:
   \[
   \approach{}_n^i(\triple{h}{r}{t}) = \mathcal{R}e\left( \int_{\Omega} \mathbf{h}(x) \mathbf{r}(x) \mathbf{t}(x) \, dx \right)
   \]
   By swapping $\mathbf{h}$ and $\mathbf{t}$:
   \begin{align}
       \approach{}_n^i(\triple{t}{r}{h}) &= \mathcal{R}e\left( \int_{\Omega} \mathbf{t}(x) \mathbf{r}(x) \mathbf{h}(x) \, dx \right)\\
       & = \mathcal{R}e\left( \int_{\Omega} \mathbf{h}(x) \mathbf{r}(x) \mathbf{t}(x) \, dx \right) \\
       &= \approach{}_n^i(\triple{h}{r}{t}).
   \end{align}

\end{proof}
  \end{itemize}

\end{document}